# Associative Memory Model with Neural Networks: Memorizing multiple images with one neuron

Hiroshi Inazawa

*Center for Education in Information Systems, Kobe Shoin Women's University[1],*
*1-2-1 Shinohara-Obanoyama, Nada Kobe 657-0015, Japan.*
E-mail: inazawa706@gmail.com

This paper presents a neural network model (associative memory model) for memory and recall of images. In this model, only a single neuron can memorize multi images and when that neuron is activated, it is possible to recall all the memorized images at the same time. The system is composed of a single cluster of numerous neurons, referred to as the "Cue Ball," and multiple neural network layers, collectively called the "Recall Net."  One of the features of this model is that several different images are stored simultaneously in one neuron, and by presenting one of the images stored in that neuron, all stored images are recalled. Furthermore, this model allows for complete recall of an image even when an incomplete image is presented

## 1. Introduction

This paper presents a neural network model for memory and recall of images. In this model, only a single neuron can store[2] multi images and when that neuron is activated, it is possible to recall all the stored images one after another. In addition, additional memories can be formed at any time, eliminating the interference between each memory. Also, even if an incomplete image pattern is presented, the complete pattern can be recalled. During recall, only a single pattern image is presented, but different pattern images that were stored at the same time as the presented pattern image are recalled and displayed on multiple recall nets. In the sense that multiple related patterns can be recalled from the presentation of a single pattern, this may be considered a form of associative memory. It could be described, so to speak, as a chain of associative memory. This behavior suggests that the memory and recall model proposed here is considered to share certain similarities with the functioning of the human brain. For example, when we recall a particular memory, it is common for other items that were stored at the same time to come to mind in rapid succession. Furthermore, it is considered that the recall threshold setting in this model can control whether a recalled memory reaches conscious awareness or not. If this threshold varies between individuals, it could contribute to individual differences in memory recall.

In other words, the model proposed here can be considered to fall within the domain of associative memory [2-5] when described in terms of conventional neural networks. In associative memory models, presenting a fragment of stored information enables the recall of the entire memory, much like how human memory processing works. Conventional associative memory models generally consist of an input layer consisting of a large number of neurons and an output layer consisting of a small number of neurons that represent the processing results (in some models, the input layer and output layer are integrated). Such associative memory models were actively researched during the second neural network boom [2-8], when the memory capacity of these models became a major problem. Both simulations and analytical studies have shown that the maximum memory capacity is approximately 15% of the number of neurons in the input layer. Exceeding this limit leads to interference among the stored memories [5-8]. As described earlier, the present study explores the possibility of a model that, in principle, allows for additional learning and the storage of multiple (related) memories, without being limited by memory capacity constraints. On the other hand, neural network-based learning has currently achieved great success [9-10]. In particular, since the 2000s, research on learning using images has become highly active, with an explosive number of studies being published [11-14]. These approaches, now known as deep learning and generative AI, have achieved significant success in the field of artificial intelligence [15-16]. Much of the research in Deep Learning is based on techniques developed during the second wave of neural network research, which took place from the 1980s to the early 1990s and focused on multi-layer neural networks [17-26]. Meanwhile, although several notable studies on memory have been published during the same period [2] [5-8] [25] [26], so far, they do not have achieved the same level of impact or success as those on learning. However, recently it has been suggested that memory can be modeled using learning methods employed in transformer theory [27]. Note that this paper is based on a submission from 2021 [1], which provides a detailed explanation of the storing and recalling of a single pattern.

---

[1] This was my affiliated institution until the end of March 2025, when I retired. Please note that as of April 2025, the university name has been changed from "Kobe Shoin Women's University" to "Kobe Shoin University."

[2] A grandmother cell has previously been proposed, i.e., where one neuron is responsible for memory.

Section 2 describes the details of the model, section 3 presents the simulation results, and section 4 is devoted to conclusions and discussion.

## 2. Specification of the model

Let's we provide an overview of the model. The proposed model performs bidirectional associative learning between a single neuron (called the cue neuron) in a cluster of many neurons called the "cue ball" and multiple neurons (called the recall neurons) in several Recall Nets. The term "cue neuron" is used here because this neuron is positioned as the trigger that initiates the recall of stored patterns. The learning is executed for multiple patterns with respect to a single cue neuron. During the recall process, after bidirectional learning between cue neurons and recall neurons, the output values generated from a single pattern image presented to the Recall Net are fed into all cue neurons within the Cue Ball. At this time, the cue neuron that has memorized the presented pattern image outputs a higher output value than the other cue neurons. Using the output value of this cue neuron, different patterns memorized in multiple Recall Nets are recalled. Let's explain this in more detail by making it a bit more schematic. The relation between the Cue Ball and the Recall Net can be schematically represented as shown in Figure 1. The overall shape of the Cue Ball can, of course, be arbitrary; however, for illustrative purposes, we have adopted a spherical form as shown in Figure 1. Next, we briefly explain the mechanism of memory and recall in this model. Note that the details have already been explained in [1]. A cue neuron is connected to all recall neurons in several Recall Nets, which in our simulation is connected to three separate Recall Nets. The cue neurons in one Cue

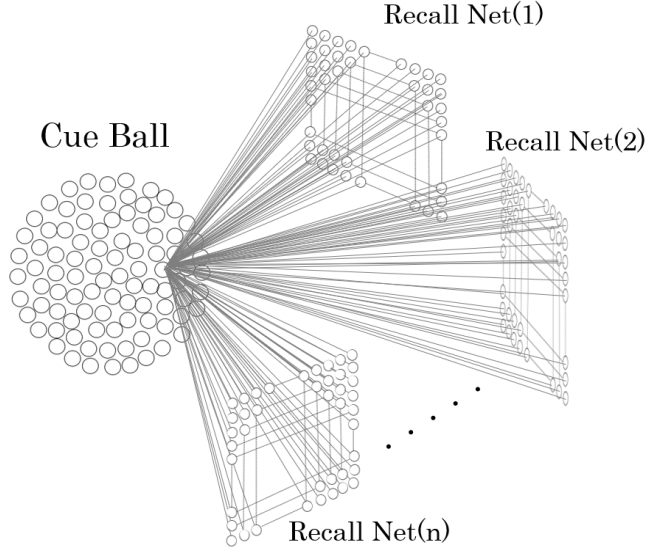

Figure 1: Schematic diagram of the Cue Ball and the Recall Net. The Cue Ball is represented by a sphere with Recall Nets around it. Every recall neuron in each Recall Net is connected to cue neurons in the Cue Ball.

Ball are connected to all of the recall neurons in multiple Recall Nets. In addition, there are no connections between cue neurons in the Cue Ball, between recall neurons in the Recall Net, or between recall neurons in other Recall Nets. Although the number of cue neurons in the cue ball can in principle be increased without limit, in this study we prepared 60,000 cue neurons to match the number of prepared pattern images. Note that the actual number of cue neurons used is 3,000 (1000 × 3). In other words, in the simulation, three Recall Nets are used, and each Recall Net memorizes different 1,000 pattern images. On the other hand, each Recall Net consists of 784 neurons, corresponding to the number of pixels in a single pattern image (28 pixels in height × 28 pixels in width). Each cue neuron is connected to all recall neurons within Recall Nets and has a threshold. There are no internal connections among cue neurons or among recall neurons.

The processing procedure consists of two processes: the storing (learning) encoding process and the recall process. First, let's explain how the learning process works. The memorizing (learning) is performed by adjusting the connection weights from a single cue neuron, selected from the Cue Ball, to all recall neurons in a Recall Net. The data to be memorized consists of a set of handwritten digit pattern images known as MNIST [9]. Each pattern is composed of 784 pixels. Note that actual patterns are represented by digital values derived from converting character patterns from pixel data into grayscale. For this reason, one Recall Net requires 784 recall neurons. After presenting a single pattern to the Recall Net, learning takes place between all recall neurons in this layer and a single cue neuron connected to them, through the adjustment of the connection weights $w_{ji}^g$ of the recall neurons. Subsequently, the outputs from all recall neurons in the Recall Net are fed back into that cue neuron, and learning occurs through the adjustment of the cue neuron's connection weights $v_{ij}^g$. This completes the bidirectional learning process between the cue neuron and the recall neurons.

Next, we describe the process of the recall mechanism. A previously learned image pattern is presented to the Recall Net. Subsequently, the output values of all recall neurons in that Recall Net are fed into all cue neurons. Each cue neuron receiving input generates an output; if a neuron outputs a value near the learned threshold $\theta$ set during learning, that cue neuron is regarded as having learned the presented pattern. By normalizing the output value of this cue neuron to 1.0 using a threshold function and feeding it back to the recall neurons of the corresponding Recall Net, the memorized pattern—identical to the originally presented one—will be reproduced in the Recall Net. If the threshold is lowered, multiple patterns similar to the originally presented character pattern emerge as recall candidates. In the following, we provide a more detailed description of the above learning and recall processes using mathematical expressions. We begin with the learning process followed by the recall process. The input-output relation between a single cue neuron and all recall neurons within a Recall Net can be described by the following equation.

$$y_j = w_{ji}^g x_i \quad , \tag{1}$$

where $x_i$ denotes the output of the i-th cue neuron, and $y_j$ denotes the output of the j-th recall neuron. $w_{ji}^g$ denotes the connection weight from the i-th cue neuron to the j-th recall neuron, and $g$ indicates the partition group of the presented character pattern image. Although this equation is commonly seen in many contexts, it should be noted that the summation over "i" is not taken in this case. The learning process employs the gradient descent method (GDM) [19] [20], under which the error function $E$ for the recall neurons is defined as follows.

$$E \equiv \frac{1}{2} \sum_{j=0}^{M} \left(d_j^p - y_j\right)^2 \quad , \tag{2}$$

where $d_j^p$ represents the j-th element of the presented character pattern, where p is the pattern index assigned across all pattern groups, and $M$ denotes the number of recall neurons within a single Recall Net ($M = 784 - 1$). The following equation describes the weight update rule obtained using the gradient descent method (GDM).

$$w_{ji}^g(t+1) = w_{ji}^g(t) + \Delta w_{ji}^g(t) \tag{3}$$

$$\Delta w_{ji}^g(t) = -\varepsilon_W \frac{\partial E}{\partial w_{ji}^g} = \varepsilon_W \left(d_j^p - y_j(t)\right) x_i(t) \quad , \tag{4}$$

where $t$ denotes the number of updates, and $\varepsilon_W$ is the learning rate, which is set to 1. By applying the input-output relation given in eq. (1) along with the weight update rules in eqs. (3) and (4), it can be shown that $w_{ji}^g$ can reliably learn the presented pattern. The following provides an analytical demonstration of this. To begin, we evaluate Eq. (1) at $t$+1 by applying the update rules defined in eqs. (3) and (4).

$$y_j(t+1) = w_{ji}^g(t+1) x_i(t+1)$$
$$= (w_{ji}^g(t) + (d_j^p - w_{ji}^g x_i(t)) x_i(t+1) \tag{5}$$

Moreover, if we assume the output $x_i(t) = x_i(t+1) \equiv 1$, eq. (5) can be rewritten as follows.

$$y_j(t+1) = d_j^p \tag{6}$$

Thus, after the learning, $y_j(t+1)$ becomes identical to the corresponding element of the presented character pattern. Next, we examine the learning process of the connection weights from recall neurons to cue neurons during input to the cue neurons. The input-output relation from recall neurons to cue neurons is given as follows.

$$x_i = f(q_i) = f\left(\sum_{j=0}^{M} v_{ij}^g y_j\right) = \begin{cases} 0 & for\ q_i < D \\ 1 & for\ q_i \geq D \end{cases}, \quad (7)$$

where $q_i$ denotes the output of the cue neuron before thresholding, and $y_j$ represents the input to the j-th recall neuron. In this case, we use the value of $y_j$ which is obtained from eq. (1) by setting $x_i=1$ after the weights $w_{ji}^g$ are learned. $v_{ij}^g$ represents the connection weight from the j-th recall neuron to the i-th cue neuron, f is the threshold function, and $D$ denotes the threshold value. Learning is carried out using the same GDM method as in the case of $w_{ji}^g$ learning. The error function for the cue neurons in this case is given as follows.

$$e \equiv \frac{1}{2}\sum_{i=0}^{all}(\theta - q_i)^2, \quad (8)$$

where "$\theta$" is an arbitrary constant value set as the learning value for $v_{ij}^g$. Using these, the weight update rule derived by the GDM method is given as follows.

$$v_{ij}^g(t'+1) = v_{ij}^g(t') + \Delta v_{ij}^g(t') \quad (9)$$

$$\Delta v_{ij}^g(t') = -\varepsilon_V \frac{\partial e}{\partial v_{ij}^g} = \varepsilon_V(\theta - q_i(t'))y_j(t'), \quad (10)$$

where $t'$ denotes the number of updates, and $\varepsilon_V$ is the learning rate. As with $\varepsilon_W$, $\varepsilon_V$ is set to 1. Similarly, to the learning of $w_{ji}^g$, by using the input-output relation of eq. (7) along with the weight update rules in eqs. (9) and (10), it is possible to reliably learn $v_{ij}^g$. where, we consider that "$y_j$" in equation (1) is equal to "$d_j^p$", and set the following normalization condition for "$d_j^p$".

$$q_i(t'+1) = \sum_{j=0}^{M} v_{ij}^g(t'+1)y_j$$

$$= \sum_{j=0}^{M} v_{ij}^g(t')y_j\left(1 - \sum_{k=0}^{M}(y_k)^2\right) + \theta \sum_{j=0}^{M}(y_j)^2, \quad (11)$$

where, we consider that "$y_j$" in equation (1) is equal to "$d_j^p$", and set the following normalization condition for "$d_j^p$".

$$\sum_{j=0}^{M}(y_j)^2 = \sum_{j=0}^{M}(d_j^p)^2 \equiv 1 \quad (12)$$

Using this normalization condition, eq. (11) becomes as follows.

$$q_i(t'+1) = \theta \quad (13)$$

In this way, $q_i(t'+1)$ matches the learning value "$\theta$" after the learning of $v_{ij}^g$.

## 3. Simulation and Results

In this section, we verify the working of the learning and recall processes described in Section 2 through simulation. As previously mentioned, the pattern data used for the simulation is handwritten digit images from the MNIST dataset. Each pattern is composed of 28 × 28 pixels (=784 pixels) [9]. In the simulation, the 8-bit pixel binary image data that make up the character patterns are converted into grayscale digital data for use. Although the total number of images is 60,000, the simulation actually uses 3,000 of these patterns. These 3,000 patterns are divided into three groups, with 1,000 patterns used per group. During learning of $w_{ji}^g$, one pattern is sequentially taken from the beginning of each pattern group, and its element values are assigned to $d_j^p$ in eq. (4). Specifically, the 0th pattern from the first group, the 1,000th pattern from the second group, and the 2,000th pattern from the third group are selected and presented to Recall Net. The simulation algorithm for the learning and recall processes is presented below. We begin by describing the learning procedure for $w_{ji}^g$.

1. Specify the pattern group "$g$":
   The following procedure is repeated sequentially for Group 1 through Group 3 (i.e., $g$ = 0, 1, 2).

2. Extract character patterns from the pattern group: End when all patterns are extracted.

3. Initialize all input/output variables and connection weights, and set the learning rate:

   $x_i = 1.0, q_i = 0.0, y_j = 1.0, w_{ji}^g = 1.0, \varepsilon_W = 1.0$

4. Select a cue neuron, apply the initial values to eq. (1), and calculate $y_j$

5. Apply the element values of the presented pattern, along with Steps 3 and 4 above, to eqs. (3), (4) to perform learning of the connection weights:

$$w_{ji}^g(t+1) = w_{ji}^g(t) + \Delta w_{ji}^g(t), \ \Delta w_{ji}^g(t) = \varepsilon_W \left(d_j^p - y_j(t)\right) x_i(t)$$

6. Save the post-learning values of $w_{ji}^g$ to a file

7. Repeat Steps 2 through 6 while switching the pattern Group

Next, the learning algorithm for $v_{ij}^g$ is summarized below.

1. Specify the pattern group "$g$":
   The following procedure is repeated sequentially for group 1 through Group 3 (i.e., $g$ = 0, 1, 2).

2. Initialize all input/output variables and connection weights, and set the learning rate:

   $x_i = 1.0, q_i = 0.0, y_j = 1.0, v_{ij}^g = 1.0, \varepsilon_V = 1.0, \theta = 100.0, D = 90.0$

3. Calculate $q_i$ of eq. (7) for all cue neurons: Using $y_j$ computed from the trained $w_{ji}^g$

4. Apply the threshold function defined in eq. (7) to $q_i$, and calculate $x_i$:

   $x_i = 1.0 \ for \ q_i \geq D \ \& \ x_i = 0.0 \ for \ q_i < D$

5. Perform learning of the connection weights using eq. (9), based on $q_i$ and $y_j$:

   $x_i = 1.0, q_i = 0.0, y_j = 1.0, v_{ij}^g = 1.0, \varepsilon_V = 1.0, \theta = 100.0, D = 90.0$

6. Save the post-learning values of $w_{ji}^g$ to a file

7. Repeat Steps 2 through 6 while switching the pattern Group

First, let's verify whether the learning proceeds as expected under these algorithms.

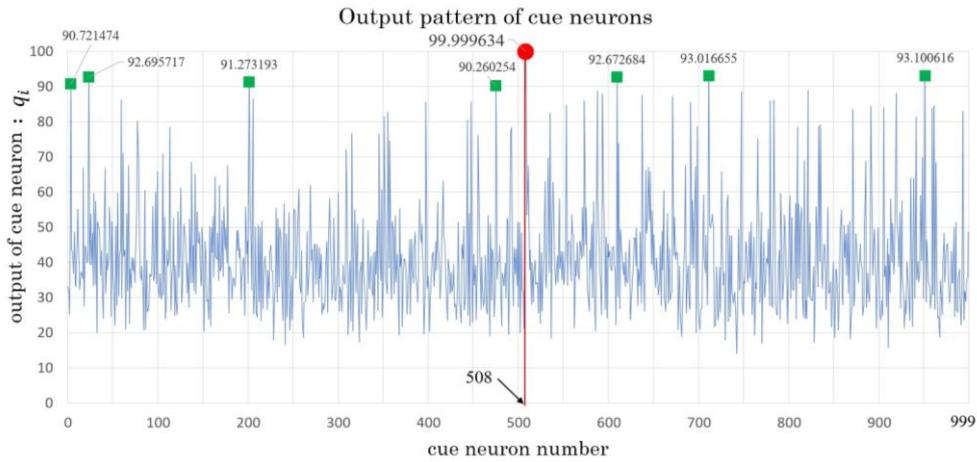

Figure 2a: Output pattern of the cue neurons. The vertical axis represents $q_i$ values for cue neurons indexed from 0 to 999. The presented learned pattern corresponds to pattern No. 508 from pattern Group 0. Accordingly, cue neuron 508 exhibits the strongest activation, with an output value of 99.99963. In addition, seven other cue neurons have been identified with output values exceeding the threshold of 90. These $q_i$ values are also annotated in the figure.

Following the above algorithm, after learning $w_{ji}^g$ and $v_{ij}^g$, the output values $q_i$ of all cue neurons (indices 0–999) in response to the learning pattern No. 508 from pattern Group 0 are shown in Figure 2a. In this figure, the maximum $q_i$ is 99.99963, produced by neuron 508, which corresponded to the trained pattern No. 508. Note that if the presented pattern has been previously learned, the maximum value of $q_i$ will be close to the learned value $\theta = 100.0$. Although significant figures should normally be considered when presenting numerical values, we omit such rounding here to clearly highlight differences between values in this paper. In Figure 2a, in addition to neuron 508, the output values of seven other neurons are significantly elevated, when the threshold $D \geq 90.0$. These seven patterns have been detected because they share many similar features with the presented pattern. Figure 2b shows the image of these patterns.

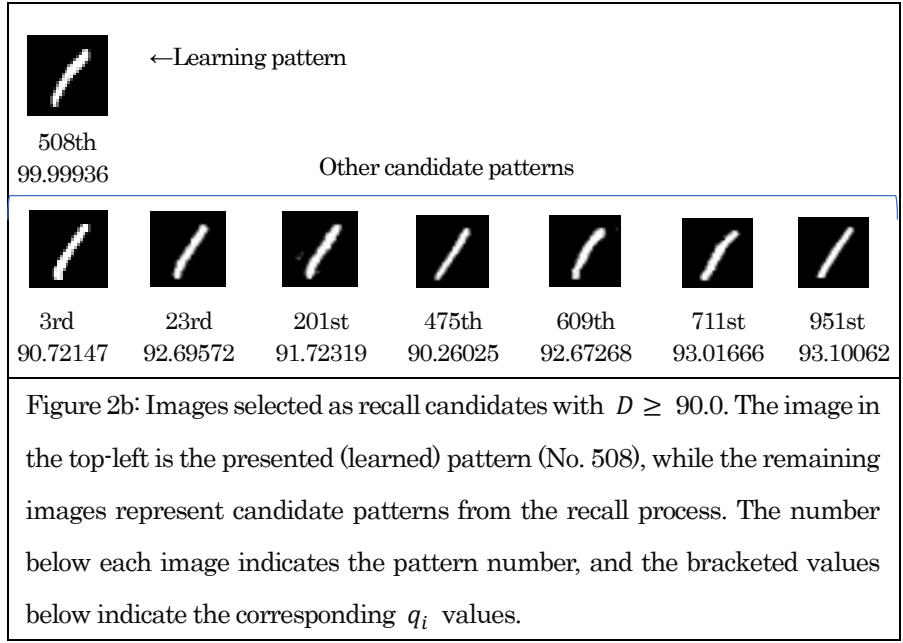

Figure 2b: Images selected as recall candidates with $D \geq 90.0$. The image in the top-left is the presented (learned) pattern (No. 508), while the remaining images represent candidate patterns from the recall process. The number below each image indicates the pattern number, and the bracketed values below indicate the corresponding $q_i$ values.

Next, the algorithm of the recall process is described below.

1. Load the learned connection weights from a file: $w_{ji}^g$、 $v_{ij}^g$ for $g = 0,1,2$
2. Specify the pattern Group and the presented pattern
3. Using the element values of the presenting (learned) pattern from pattern Group 0 ($g = 0$), compute the output value $q_i$ for all cue neurons, and identify the cue neuron with the largest $q_i$
4. Using a threshold, compute $x_i$ for the output $q_i$ of the cue neuron identified in step 3:

    $x_i = 1.0$ for $q_i \geq D$ & $x_i = 0.0$ for $q_i < D$

5. Using the output value $x_i = 1.0$ of this cue neuron, calculate $y_j$ using eq. (1).
6. Display the output pattern using the output values $y_j$
7. Repeat steps 5 and 6 for the remaining pattern Groups ($g = 1,2$)

Let's now execute this recall algorithm. For the presented pattern group and the specific (learning) pattern to be shown, we use pattern No. 508 from Group 0, as illustrated in Figure 2a. In this case, the index of the cue neuron used for learning is identical to the pattern number presented. If learning has been successful, the activation of this cue neuron should naturally trigger the recall of pattern No. 508 from the presented pattern group, No. 1508 from Group 1, and No. 2508 from Group 2. Figure 3 shows the result of this recall process.

Activation cue neuron number=508th

```
0 0 0 0 0 0 0 0 0 0 0 0 0 0 0 0 0 0 0 0 0 0 0 0 0 0 0 0
0 0 0 0 0 0 0 0 0 0 0 0 0 0 0 0 0 0 0 0 0 0 0 0 0 0 0 0
0 0 0 0 0 0 0 0 0 0 0 0 0 0 0 0 0 0 0 0 0 0 0 0 0 0 0 0
0 0 0 0 0 0 0 0 0 0 0 0 0 0 0 0 0 0 0 0 0 0 0 0 0 0 0 0
0 0 0 0 0 0 0 0 0 0 0 0 0 0 0 0 0 0 0 0 0 0 0 0 0 0 0 0
0 0 0 0 0 0 0 0 0 0 0 0 0 0 0 0 0 7 171 218 36 0 0 0 0 0 0 0
0 0 0 0 0 0 0 0 0 0 0 0 0 0 0 0 12 60 253 253 91 0 0 0 0 0 0 0
0 0 0 0 0 0 0 0 0 0 0 0 0 0 0 0 70 253 253 253 91 0 0 0 0 0 0 0
0 0 0 0 0 0 0 0 0 0 0 0 0 0 0 8 94 232 253 253 232 59 0 0 0 0 0 0
0 0 0 0 0 0 0 0 0 0 0 0 0 0 0 47 253 253 253 236 68 0 0 0 0 0 0 0
0 0 0 0 0 0 0 0 0 0 0 0 0 0 0 128 219 253 253 253 137 0 0 0 0 0 0 0
0 0 0 0 0 0 0 0 0 0 0 0 0 0 45 239 253 253 230 160 8 0 0 0 0 0 0 0
0 0 0 0 0 0 0 0 0 0 0 0 0 4 146 253 253 253 166 0 0 0 0 0 0 0 0 0
0 0 0 0 0 0 0 0 0 0 0 0 0 112 254 253 253 190 15 0 0 0 0 0 0 0 0 0
0 0 0 0 0 0 0 0 0 0 0 0 0 185 253 253 232 12 0 0 0 0 0 0 0 0 0 0
0 0 0 0 0 0 0 0 0 0 0 0 74 233 253 199 70 0 0 0 0 0 0 0 0 0 0 0
0 0 0 0 0 0 0 0 0 0 0 0 68 246 253 253 101 0 0 0 0 0 0 0 0 0 0 0
0 0 0 0 0 0 0 0 0 0 0 0 139 253 253 218 15 0 0 0 0 0 0 0 0 0 0 0
0 0 0 0 0 0 0 0 0 0 0 26 216 253 223 40 0 0 0 0 0 0 0 0 0 0 0 0
0 0 0 0 0 0 0 0 0 0 0 109 253 253 68 0 0 0 0 0 0 0 0 0 0 0 0 0
0 0 0 0 0 0 0 0 0 0 0 22 208 253 233 52 0 0 0 0 0 0 0 0 0 0 0 0
0 0 0 0 0 0 0 0 0 0 0 94 253 253 70 0 0 0 0 0 0 0 0 0 0 0 0 0
0 0 0 0 0 0 0 0 0 0 0 94 253 194 8 0 0 0 0 0 0 0 0 0 0 0 0 0
0 0 0 0 0 0 0 0 0 0 0 94 253 115 0 0 0 0 0 0 0 0 0 0 0 0 0 0
0 0 0 0 0 0 0 0 0 0 0 36 218 115 0 0 0 0 0 0 0 0 0 0 0 0 0 0
0 0 0 0 0 0 0 0 0 0 0 0 0 0 0 0 0 0 0 0 0 0 0 0 0 0 0 0
0 0 0 0 0 0 0 0 0 0 0 0 0 0 0 0 0 0 0 0 0 0 0 0 0 0 0 0
0 0 0 0 0 0 0 0 0 0 0 0 0 0 0 0 0 0 0 0 0 0 0 0 0 0 0 0
```

$g$=0, presented pattern=508th

```
(g=1 pattern matrix)
```

$g$=1, pattern=1508th

```
(g=2 pattern matrix)
```

$g$=2, pattern=2508th

Figure 3: Three recalled patterns resulting from the activation of cue neuron No. 508. This figure displays the digital data of the recall patterns in each pattern group. The top panel shows the recalled pattern of pattern Group 0, the middle panel shows the recalled pattern from Group 1, and the bottom panel shows the recalled string pattern from the Group 2.

Note that this figure displays the patterns in terms of their digital values, representing the state of the Recall Net. The top panel shows the output from Group 0 when cue neuron No. 508 is activated. The middle and bottom panels show the Recall Net's output patterns from Group 1 and Group 2, respectively, as learned through this cue neuron (pattern No. 1508 in the middle panel, and No. 2508 in the bottom panel). The corresponding pixel-based representations of these patterns are shown in Figure 4.

Activation cue neuron number=508th

| Recall Net (1) | Recall Net (2) | Recall Net (3) |
| --- | --- | --- |
| pattern=508th | pattern=1508th | pattern=2508th |

Figure 4: The images of three output patterns recalled by cue neuron No. 508. The patterns are shown as pixel-based representations (image) of the recalled patterns from each pattern group. The top panel shows the recalled pattern from Group 0, the middle panel shows the recalled pattern from Group 1, and the bottom panel shows the recalled pattern from Group 2.

As shown, when three image patterns are learned in association with a single cue neuron, presenting one of the patterns trigger the recall of the other two patterns learned at the same time. Now, let's examine the case in which the initially presented learned pattern is incomplete. We examine the case in which the lower half of pattern No. 508 is not presented. When this incomplete pattern is presented to the Recall Net, if the corresponding cue neuron exhibits the largest output value $q_i$ among all cue neurons, the resulting output behavior is the same as that shown in Figure 4. This behavior is illustrated in Figure 5.

Recall Net (1), presented pattern of upper half of 508th

Figure 5: Presenting the top 50% of pattern number 508 in Pattern Group 0. In this case, the cue neuron corresponding to pattern No. 508 exhibits the largest $q_i$ value, which is 53.96446.

Among the 1,000 cue neurons, the one with the largest output value is again neuron No. 508. Its $q_i$ value in this case is 53.96446. This value is smaller than the threshold $D$, reflecting the fact that the input signal from the Recall Net is weak. In particular, the results obtained here have been consistent whether the patterns used for presentation (learning) is varied or the number of patterns studied simultaneously is the same.

## 4. Conclusions and Discussions

In this paper, we have proposed a neural network model that can memorize (learn) and recall handwritten character images. This system consists of a mass of many neurons called a "Cue Ball" and multiple neural network layers called "Recall Net." In this system, a single cue neuron engages in bidirectional learning with many different recall neurons distributed across multiple Recall Net layers. After learning, when one of the learned images is presented to the Recall Net corresponding to the cue neuron that memorized it, all images associated with that cue neuron are recalled across the corresponding Recall Net. This behavior may resemble certain aspects of how memory and recall function in the human brain. For example, when we recall a particular memory, it is common for other (memory) information that was learned simultaneously to come to mind as well. In the sense that multiple related patterns can be recalled from a single presented pattern, this can be regarded as a form of associative memory. Furthermore, the strength of recall can be modulated by the threshold value used in this model. This may be interpreted as a mechanism that controls whether a memory enters conscious awareness or not. If this threshold varies between individuals, it might be account for individual differences in recall characteristics. Furthermore, in the simulation, we also have checked the case where only half of a stored pattern was presented. Even in this case, the cue neuron associated with the original memory could be identified, and using that neuron, all the corresponding memory patterns have been displayed in the Recall Net. This can be thought of as similar to how our brains reconstruct complete memories from partial or fading fragments.

As previously explained, this model memorizes images via a learning process. During training, data is stored in the form of connection weights, and the space required for this data grows linearly with the number of images processed. Let's calculate the total amount of connection weight data generated in this simulation. There are two types of connection weight parameters: $w_{ji}^g$ and $v_{ij}^g$. The index gg takes three values ($g = 0, 1, 2$). The index i ranges over the cue neurons ($i = 0$ to 999), and j corresponds to the recall neurons in a single Recall Net image (j = 0 to 783). Therefore, the total number of parameters for $w_{ji}^g$ is $g \times j \times i = 3 \times 784 \times 1{,}000 = 2{,}352{,}000$. The situation is the same for $v_{ij}^g$. Considering the data size of one connection weight, the total memory required for $w_{ji}^g$ is about 18Mbyte. Including the weights for $v_{ij}^g$, the total is about 36Mbyte. Thus, the memory usage in this simulation is approximately 18 MB. Therefore, as the number of images to be stored increases in the future, it will likely become necessary to take measures to reduce memory consumption, such as increasing the hardware's memory capacity. It is considered that a model that can recall multiple related memories is an interesting direction for future research.

## References

[1]  H. Inazawa, "Hetero-Correlation-Associative Memory with Trigger Neurons: Accumulation of Memory through Additional Learning in Neural Networks," Complex Systems, **27**, Issue 2, 2018, pp. 187-197.;
H. Inazawa, "An associative memory model with very high memory rate: Image storage by sequential addition learning," 2022 arXiv: **2210.03893**, https://doi.org/10.48550/arXiv.2210.03893.

[2]  T. Kohonen, "Correlation Matrix Memories," *IEEE Trans.*, **C-21**, 1972 pp. 353-359.;
T. Kohonen, "*Self-Organization and Assocoative Memory,*" Springer, 1984.

[3]  J.R. Anderson and G.H. Bower, "*Human Associative Memory,*" Psychology Press, 1974.

[4]  K. Nakano, "*Learning Process in a Model of Associative Memory,*" Springer, 1971.

[5]  S. Amari and K. Maginu, "Statistical Neurodynamics of Associative Memory. *Neural Networks,*" **Vol. 1**, 1988 pp. 63-73.

[6]  J.J. Hopfield, "Neural Networks and Physical Systems with Emergent Collective Computational Abilities," *Proc. of the National Academy of Science USA* **79** : 1982 pp. 2254-2258.
J.J. Hopfield and D.W. Tank, "Neural computation of decisions in optimization problems," Biol Cybern **52**, 1985 pp.141–152.

[7]  S. Yoshizawa, M. Morita and S. Amari, "Capacity of Associative Memory Using a Nonmonotonic Neuron Model," *Neural Networks,* **Vol. 6**, 1993 pp. 167-176.

[8]  D.J. Amit, H. Gutfreund and H. Sompolinsky, "Storing Infinite Numbers of Patterns in a Spin-Glass Model of Neural Networks," *Phys. Rev. Lett.,* **55**, 1985 pp. 1530-1533.

[9]  G.E. Hinton and R. Salakhutdinov, "Reducing the Dimensionality of Data with Neural Networks," *Science,* **Vol. 313**, 2006 pp. 504-507.

[10]  G.E. Hinton, S. Osindero and Y. Teh, "A Fast Learning Algorithm for Deep Belief Nets," *Neural Computation,* **18**, 2006 pp. 1527-1544.

[11]  Y. Bengio, P. Lamblin, D. Popovici and H. Larochelle, "Greedy Layer-Wise Training of Deep Networks," *In Proc. NIPS,* 2006.

[12]  H. Lee, R. Grosse, R. Ranganath and A.Y. Ng, "Convolutional Deep Brief Networks for Scalable Unsupervised Learning of Hierarchical Representations," *In Proc. ICML,* 2009.

[13]  A. Krizhevsky, I. Sutskever and G.E. Hinton, "ImageNet Classification with Deep Convolutional Neural Networks," *In Proc. NIPS,* 2012.

[14]  Q.V. Le, M. Ranzato, R. Monga, M. Devin, K. Chen, G.S. Corrado, J. Dean and A.Y. Ng, "Building High-Level Features Using Large Scale Unsupervised Learning," *In Proc. ICML,* 2012.

[15]  A. Vaswani, N. Shazeer, N. Parmar, J. Uszkoreit, L. Jones, A.N. Gomez, L. Kaiser and I. Polosukhin, "Attention Is All You Need," 2017.

[16]  J.W. Rae, A. Potapenko, S.M. Jayakumar and T. Lillicrap, "*Compressive Transformers for Long-Range Sequence Modeling,*" 2020.

[17]  D.E. Rumelhart, J.L. McCleland and the PDP Research Group, "Parallel Distributed Processing:

*Explorations in the Microstructure of Cognition,"* **Vol. 1**: *Foundations. Cambridge: MIT Press,* 1986.

[18] K. Fukushima and S. Miyake, "Neocognitron: A new Algorithm for Pattern Recognition Tolerant of Deformations and Shifts in Position," *Pattern Recognition,* **15**: 1982 pp. 455-469.

[19] B. Widrow and M.E. Hoff, "Adaptive Switching Circuits," *In 1960 IRE WESCON Convention Record, Part 4,* 96-104. *New York: IRE. Reprinted in Anderson and Rosenfeld [1988],* 1960.

[20] R.A. Rescorla and A.R. Wagner, "A Theory of Pavlovian Conditioning: The Effectiveness of Reinforcement and Nonreinforcement," *In classical Conditioning II: Current Research and Theory, eds. A.H.* Bloch and W.F. Prokasy, *New York: Appleton-Century-Crofts,* 1972 pp. 64-69.

[21] Y. LeCun, B. Boser, J.S. Denker, D. Henderson, R.E. Howard, W. Hubard and L.D. Jackel, "Backpropagation Applied to Handwritten Zip Code Recognition. Neural Computation," **1(4)**, 1989 pp. 541-551.

[22] K. Tsutsumi, "Cross-Coupled Hopfield Nets via Generalized-Delta-Rule-Based internetworks," *In Proc. IJCNN90-San Diego II,* 1990 pp. 259-265.

[23] P.Y. Simard, D. Steinkraus and J. Platt, "Best Practice for Convolutional Neural Networks Applied to Visual Document Analysis," *In Proc. ICDAR,* 2003.

[24] Y. LeCun, L. Bottou, Y. Bengio and P. Haffner, "Gradient-Based Learning Applied to Document Recognition," *Proceedings of the IEEE,* **86(11)**: 1998 pp. 2278-2324.

[25] N. Srebro, and A. Shraibman, "Rank, Trace-Norm and Max-Norm," *In Proc. 18th Annual Conference on Learning Theory, COLT 2005,* Springer, 2005 pp. 545-560.

[26] N. Srivastava, G.E. Hinton, A. Krizhevsky, I. Sutskever, and R. Salakhutdinov, "Dropout: A Simple Way to Prevent Neural Networks from Overfitting. Journal of Machine Learning Research," **15**: 2014 pp. 1929-1958.

[27] M. Cornia, M. Stefanini, L. Baraldi, and R. Cucchiara, "Meshed-Memory Transformer for Image Captioning," arXiv:**1912.08226**, 2020 https://doi.org/10.48550/arXiv.1912.08226